\documentclass[preprint,12pt]{elsarticle}

\usepackage{url}
\usepackage{amssymb}
\usepackage{subcaption}
\usepackage{graphicx}  

\usepackage{amsmath}

\journal{Neurocomputing}

\begin{document}

\begin{frontmatter}



\title{Impact of Leakage on Data Harmonization in Machine Learning Pipelines in Class Imbalance Across Sites.
}


\author[1,2,3,*]{Nicolás Nieto}

\author[1,2]{Simon B. Eickhoff}
\author[3,4]{Christian Jung}

\author[5,6]{Martin Reuter}
\author[5]{Kersten Diers}

\author[+]{for the Alzheimer’s Disease Neuroimaging Initiative}

\author[3,4]{Malte Kelm}
\author[7]{Artur Lichtenberg}

\author[1,2]{Federico Raimondo}
\author[1,2]{Kaustubh R. Patil}

\affiliation[1]{Institute of Neuroscience and Medicine (INM-7: Brain and Behaviour), Research Centre Jülich, Jülich, Germany}
\affiliation[2]{Institute of Systems Neuroscience, Heinrich Heine University Düsseldorf, Düsseldorf, Germany}
\affiliation[3]{Department of Cardiology, Pulmonology and Vascular Medicine, University Hospital and Medical Faculty, Heinrich-Heine University, Duesseldorf, Germany}
\affiliation[4]{Cardiovascular Research Institute Düsseldorf (CARID), Medical Faculty, Heinrich-Heine University, Duesseldorf, Germany}
\affiliation[5]{Artificial Intelligence in Medical Imaging, German Center for Neurodegenerative Diseases (DZNE), Bonn, Germany}
\affiliation[6]{Department of Radiology, Harvard Medical School, Boston, MA, USA}
\affiliation[7]{Department of Cardiac Surgery, University Hospital and Medical Faculty, Heinrich-Heine University, Duesseldorf, Germany}

\begin{abstract}
Due to the cost and complexity of data collection in biomedical domains, it is common practice to combine data sets from multiple sites to obtain large datasets required for machine learning. However, undesired site-specific variability presents challenges. Data harmonization aims to address this issue by removing site-specific variance while preserving biologically relevant information. We show that the widely used ComBat-based harmonization improvements are driven by data leakage due to illicit use of target information when class labels are imbalanced across sites, a common scenario in biomedical domains. We propose a novel approach, PrettYharmonize, which leverages subtle differences in data harmonized using different pretended target values. Using controlled benchmark datasets and real-world magnetic resonance imaging and clinical data, we demonstrate that our leakage-free PrettYharmonize method achieves performance comparable to leakage-prone methods. As such, PrettYharmonize is a viable method to combine data sets for machine learning applications.
\end{abstract}



\begin{keyword}
Data Harmonization \sep ComBat \sep Data Leakage \sep Machine learning \sep Medical Imaging \sep Magnetic Resonance Imaging \sep medical AI \sep clinical \sep ICU

\end{keyword}

\end{frontmatter}



\section*{Introduction}
Many research fields have greatly benefited from machine learning (ML) approaches.
ML relies on large datasets to learn generalizable models, as these datasets help capture robust underlying patterns.
This makes combining multiple datasets particularly attractive, especially in domains where obtaining data from a single location is challenging.
However, combining datasets remains a significant challenge, as datasets obtained from different locations, even with similar acquisition parameters, often contain variability unrelated to relevant biological information \cite{chen2014exploration, bayer2022site, botvinik2023reproducibility}.

This undesired site-related variability may stem from systematic differences, which can be corrected, or random variations, which cannot be modeled or corrected.
The systematic variability, known as Effects of Site (EoS), is prevalent in many biomedical domains and can lead to biased results if not properly addressed \cite{solanes2023mareos}.
For example, clinical data are influenced by the acquisition site, as different hospitals use varying laboratory equipment, procedures, and criteria. 
Similarly, medical imaging data are affected by factors such as acquisition protocols, scanner drifts, and even the time of day \cite{solanes2023mareos,hu2023review}.
Magnetic Resonance Imaging (MRI) is particularly susceptible to EoS, as variability can arise from differences in magnetic field strength, room temperature fluctuations, or electromagnetic noise—even when using scanners from the same manufacturer with identical parameters \cite{li2020denoising, wachinger2021detect}. 
To address this issue, numerous Methods Aiming to Remove the Effects of Site (MAREoS) have been proposed \cite{hu2023review, da2020review, abbasi2024deep}.
MAREoS are typically employed as a preprocessing step to generate ``site-effect-free" data while preserving biological information, resulting in harmonized data that can enhance subsequent statistical and ML analyses \cite{fortin2018harmonization, acquitter2022radiomics, li2021impact, ingalhalikar2021functional,maikusa2021comparison}.

ComBat-based MAREoS are extensively used in several domains.
Originally proposed to correct batch differences in genomic data \cite{johnson2007adjusting}, ComBat was later adapted to other domains, including MRI data \cite{fortin2017harmonization, fortin2018harmonization}.
The method employs Bayesian regression to estimate additive (location) and multiplicative (scale) corrections for each feature across sites.
In addition to removing site effects, ComBat can preserve the variance of biologically relevant variables when provided as covariates.
The ComBat-GAM extension was developed to account for nonlinear covariate effects, and its associated “neuroHarmonize” software has been widely adopted \cite{pomponio2019harmonization}.
Although concerns have been raised that the assumptions of ComBat, originally designed for genomic data, may not always hold for other data types \cite{ibrahim2021radiomics}, numerous empirical studies have demonstrated its effectiveness \cite{hu2023review, da2020review, yu2018statistical, dudley2023abcd_harmonizer}.

Several studies using ComBat within ML pipelines have failed to separate training and test data, incorrectly harmonizing the entire dataset before downstream analysis \cite{fortin2017harmonization, fortin2018harmonization, acquitter2022radiomics, bourbonne2021development, campello2022minimising, castaldo2022framework, chen2023four, bostami2022multi}.
While this approach is valid for statistical analysis, it is inconsistent with ML principles, where data leakage can lead to invalid models and misleading interpretations if training and test data are not properly separated \cite{sasse2023leakage,kapoor2023leakage,lones2021avoid}.
Failing to maintain this separation during harmonization can result in overly optimistic generalization performance estimates, particularly in cross-validation settings \cite{marzi2024efficacy}.
To address this, techniques have been developed to integrate ComBat into ML pipelines as a preprocessing step, where its parameters are learned on the training data and then applied to unseen test data \cite{pomponio2019harmonization, radua2020increased, marzi2024efficacy}.

Although these methods correctly split train and test folds, further challenges arise due to the violation of a critical ComBat assumption - all variance not shared across sites is considered unwanted site-related variance.
This becomes particularly problematic when biologically relevant variance is associated with site differences, such as in a diagnostic classification task where one site predominantly includes patients and another healthy controls \cite{nygaard2016methods}, leading to class imbalance across sites.
ComBat may eliminate variance related to the target in such cases of site-target dependence, resulting in a harmonized dataset that yields fewer or null findings in subsequent analyses.
To mitigate this issue, the target variable is often included as a covariate in ComBat, ensuring its variance is preserved during harmonization.
However, while preserving relevant variance is not inherently problematic, specifying the target as a covariate introduces another form of data leakage, as the target values of the test set are required to apply the harmonization model but are unavailable in real-world scenarios \cite{sasse2023leakage}.
This critical issue has not been thoroughly investigated until recently \cite{marzi2024efficacy}.

In this work, we aimed to empirically demonstrate a limitation of ComBat-based harmonization in site-target dependence scenarios.
To this end, we conducted controlled experiments for age regression and sex classification using real MRI data from healthy individuals.
Additionally, we performed two clinically relevant tasks: dementia and mild cognitive impairment (MCI) detection using MRI data, and outcome prediction for septic patients using intensive care unit (ICU) data.
To systematically evaluate the impact of site-target association, all experiments were conducted under both site-target dependence and independence conditions.
Specifically, we investigated the performance of different harmonization schemes, both with and without preserving target variance, under these scenarios. 
Preserving target variance leads to data leakage; however, this approach is still commonly used in the literature.

Finally, to address the data leakage issue in ML pipelines while combining data from multiple sites, we propose a novel method called ``PRETended Target Y Harmonize" (\emph{PrettYHarmonize}).
Using a stacking ensemble model, PrettYHarmonize learns subtle differences in data harmonized with varying target values.
It avoids data leakage by employing pretended target values for the test data.
We validated our method using benchmark datasets \cite{solanes2023mareos} and demonstrate that \emph{PrettYHarmonize} performs competitively compared to other harmonization schemes in both site-target dependence and independence scenarios.
We provide a comprehensive comparison of no-harmonization, leakage, and no-leakage methods.
The corresponding Python package of the proposed method is publicly available at \url{https://github.com/juaml/PrettYharmonize}, and the code to replicate our results is also publicly available at \url{https://github.com/juaml/harmonize_project}.

\section*{Results}
\subsection{PrettYharmonize validation}

The proposed PrettYharmonize method builds on the neuroHarmonize model \cite{pomponio2019harmonization} but introduces a novel approach by combining two machine learning models: a \emph{Predictive} model and a \emph{Stack} model.
To harmonize test data without access to true labels, the method employs \emph{pretended} test target values.
These pretended values are used to harmonize the test data and generate predictions via the Predictive model. 
This ensures harmonization without using test labels, thereby preventing data leakage by design.

Our key assumption is that when the test data is harmonized with the correct label (i.e., when the pretended label matches the true test label), the neuroHarmonize model will preserve biologically relevant information.
Conversely, when the pretended label does not match the true label, both site-related effects and biological signals will be removed, as the harmonization is performed under an incorrect label assumption.
The Predictive model generates predictions for each harmonized dataset, one for each class in the dataset, and the Stack model combines these predictions to produce a final, unified prediction for each sample.
Since test labels are only pretended and not used, predictions can be generated without requiring access to the true test target values.
Importantly, while incorporating a novel harmonization scheme internally, PrettYharmonize only uses the harmonized data internally and does not create a harmonized dataset as outcome, but rather a final target prediction.
A detailed description of the method’s workflow is provided in the Method-PrettYharmonize section.

To evaluate the harmonization capabilities of the harmonization scheme using pretended target of PrettYharmonize, a datasets specifically designed to benchmark MAREoS was used \cite{solanes2023mareos}.
These datasets simulate eighteen MRI features, including cortical thickness, cortical surface area, and subcortical volumes, across eight internal datasets.
Among these, four datasets contain a ``True" signal, while the remaining four contain only an ``EoS" signal related to a binary target.
The EoS signal is designed to test whether MAREoS can remove it, thereby preventing fraudulent classification performance.
For each type of signal (True and EoS), two variations are provided: ``Simple" and ``Interaction" datasets, representing linear and non-linear relationships between the features and the target, respectively.
Each internal dataset comprises 1000 samples simulated from eight different sites.
A detailed description of these datasets is available in the Data-MAREoS section.

On datasets containing the True signal and no EoS, a Baseline model (Random Forest) trained on unharmonized data achieved a balanced accuracy (bACC) of approximately 80\%, as expected.
However, the same model also achieved a bACC close to 80\% on datasets containing only the EoS signal, fraudulently leveraging the EoS signal for classification (Table \ref{table:juharmonize_performance}).

Using Random Forest as both the Predictive and Stack models, PrettYharmonize successfully removed the EoS signal in all datasets containing only EoS. Furthermore, in datasets where only the True signal was present, the method preserved the real signal while aiming to remove EoS (Table \ref{table:juharmonize_performance}). To ensure robustness, we repeated this analysis using three additional Predictive models: Gaussian Process Classifier (GP), Support Vector Machine with a Radial basis kernel (SVM), and Least Absolute Shrinkage and Selection Operator (LASSO). These models yielded similar results (Tables Supp 1, 2, and 3).

\begin{table}[ht]

\centering

\caption{PrettYharmonize and Baseline (RF model without harmonization) performance on the MAREoS dataset (bACC [\%]: mean of 10 folds).}
\resizebox{1\textwidth}{!}{%

\begin{tabular}{|l|c|c|c|c|}
\hline
\textbf{Dataset Name}                & \textbf{Baseline} & \textbf{PrettYharmonize} & \textbf{Expected}& \textbf{Difference} \\ \hline
True Simple 1 (no site effect)       & 72.86          & 72.07       & As Baseline       & 0.79        \\ \hline
True Simple 2 (no site effect)       & 82.72          & 82.86       & As Baseline       & 0.06            \\ \hline
True Interaction 1 (no site effect)  & 79.43          & 79.46       & As Baseline       & 0.03            \\ \hline
True Interaction 2 (no site effect)  & 72.23          & 70.72       & As Baseline       & 1.51            \\ \hline
EoS Simple 1 (no real effect)        & 76.11          & 54.18       & Chance (50)       & 5.18            \\ \hline
EoS Simple 2 (no real effect)        & 75.35          & 52.35       & Chance (50)       & 2.35            \\ \hline
EoS Interaction 1 (no real effect)   & 77.48          & 56.20       & Chance (50)       & 6.2            \\ \hline
EoS Interaction 2 (no real effect)   & 82.79          & 58.81       & Chance (50)       & 8.81            \\ \hline
\end{tabular}}

\label{table:juharmonize_performance}
\end{table}

\subsection{Forced site-target dependence and independence scenarios.}

To investigate the impact of harmonization in scenarios where the site and target are either dependent or independent, we systematically generated different experimental conditions by sampling data from various datasets.

For site-target dependence, scenarios were created such that, for each site, the majority of samples were retained from one class, while only a small number of samples were retained from the other class(es).
For example, in a binary classification problem with two sites, site A predominantly retained samples from class one, with only a few samples from class two, while site B mainly retained samples from class two, with a small number from class one.
The inclusion of minority class samples was necessary to avoid singular matrices, which would otherwise cause computational issues for the algorithms.
We hypothesize that traditional harmonization methods will remove important biological variance in this scenario unless test labels are leaked to the harmonization model, as the biological variance is intrinsically tied to the site distribution.

In contrast, for site-target independence, scenarios were generated by retaining an equal proportion of samples from each class across all sites. In this case, it is hypothesized that harmonization will not remove biologically relevant information, even when the target is not explicitly preserved as a covariate, as the biological variance is shared across all sites.

A total of seven datasets were utilized in the experiments.
Five MRI datasets (AOMIC-ID1000, eNKI, CamCAN, SALD, and 1000Brains), containing data from healthy control participants, were sampled for age regression and sex classification tasks.
Additionally, the Alzheimer’s Disease Neuroimaging Initiative (ADNI) dataset \cite{jack2008alzheimer}, which includes data collected from multiple sites, was used to detect mild cognitive impairment (MCI) or dementia patients.
Finally, the publicly available eICU clinical dataset \cite{pollard2018eicu,pollard2019eicu} was employed to classify the hospital discharge status of septic patients, a well-studied and clinically significant problem in the literature \cite{yang2023predicting, wu2021artificial, zhang2023machine}.
A comprehensive and detailed description of all datasets, as well as the sampling methods used to enforce site-target dependence and independence, is provided in the Data Description section.

Five harmonization schemes were evaluated in this study. 
Whole Data Harmonization (\emph{WDH}) involves training a neuroHarmonize model on the pooled data from all sites to generate a harmonized dataset before splitting the data into training and test folds.
This approach inherently leads to data leakage, as the test data is included in the harmonization training process \cite{kapoor2023leakage, marzi2024efficacy}.
Another scheme, Test Target Leakage (\emph{TTL}), trains the neuroHarmonize model on the training data while explicitly retaining target variance.
Although this scheme does not use the test target to train the harmonization model, it requires test labels to transform (harmonize) the test data, resulting in data leakage.
A third scheme, \emph{No Target}, trains the harmonization model on the training data without explicitly retaining target variance.
As a result, test labels are not used for harmonizing the test set, avoiding leakage but potentially removing biologically relevant information.
A \emph{Unharmonized} approach was implemented as a baseline, where the original data  was pooled and used without any harmonization.
The proposed \emph{PrettYharmonize} model, which internally performs leakage-free harmonization, was benchmaked.
By comparing these schemes, we aimed to evaluate the trade-offs between preserving biological signals, avoiding data leakage, and maintaining model performance.

\subsubsection{Age prediction}
For the site-target dependence scenario, 200 images were extracted from each of four MRI datasets containing healthy participants (AOMIC, eNKI, CamCAN, and 1000Brains) in disjoint age ranges, thereby forcing a site-target dependence.
The same proportion of male and female participants was retained in each age range.
The \emph{Unharmonized} method achieved a Mean Absolute Error (MAE) of 6.20 (Table \ref{table:harmonization_comparison}), which falls within the expected range according to the literature \cite{more2023brain}.
Predictions using the \emph{WDH} and \emph{TTL} schemes showed an improvement in performance of approximately 2 years compared to the \emph{Unharmonized} scheme (Table \ref{table:harmonization_comparison}).

\begin{figure}
  \centering
  \begin{subfigure}{.94\textwidth}
    \includegraphics[width=\linewidth]{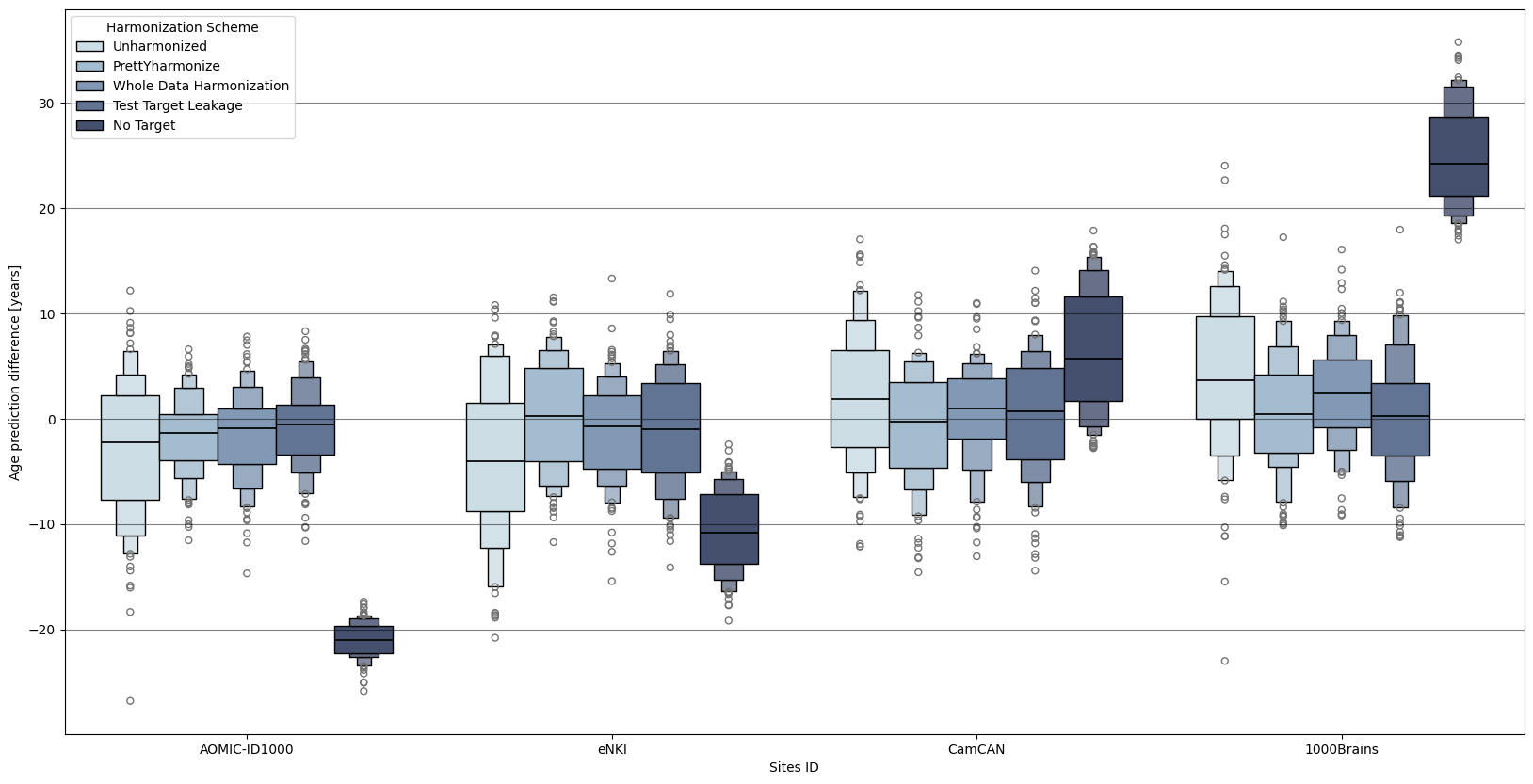}
    \caption{Age regression on site-target dependence scenario.}
    \label{fig:age_regression_desagregated_dependance}
  \begin{subfigure}{1\textwidth}
    \includegraphics[width=\linewidth]{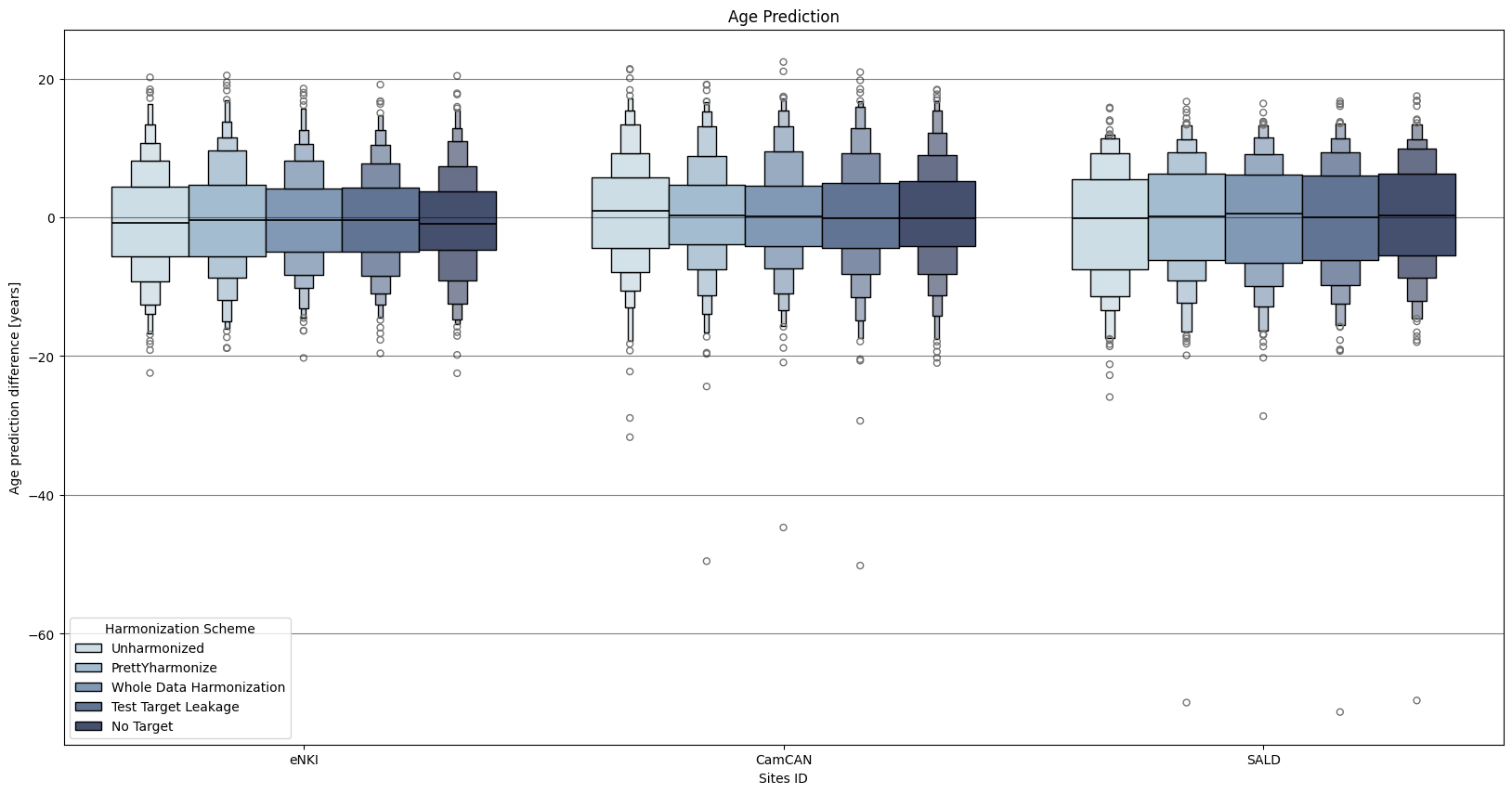}
    \caption{Age regression on site-target independence scenario}
    \label{fig:age_regression_desagregated_independance}
  \end{subfigure}
  \end{subfigure}

\caption{Age regression 
\textbf{a}) Site desegregated performance in site-target dependence scenarios.
\textbf{b}) Site desegregated performance in site-target independence scenarios.}
\end{figure}

As expected, \emph{No Target} removed the age-related signal in the features, preventing the ML model from learning the feature-target relationship and resulting in inaccurate predictions.
Specifically, the model predicted the mean population age for all individuals, leading to overestimations in the AOMIC and eNKI datasets and underestimations in the CamCAN and 1000Brains datasets (Figure \ref{fig:age_regression_desagregated_dependance}).

\emph{PrettYharmonize}, on the other hand, achieved better predictions on average compared to both the \emph{Unharmonized} and \emph{No Target} methods.
It improved the MAE, R², and age bias (Pearson's correlation between the true age and the difference between the predicted and true age) without inducing data leakage (Table \ref{table:harmonization_comparison}). Notably, \emph{PrettYharmonize}’s performance was comparable to the two leakage-prone methods (\emph{WDH} and \emph{TTL}).
Detailed individual predictions can be found in Supplementary Information Figures Supp 1-5.

For the site-target independence scenario, three datasets (eNKI, CamCAN, and SALD) containing healthy controls were used. The same number of images from each dataset was retained within the 18-80 age range, maintaining an equal proportion of males and females.
The model that used \emph{Unharmonized} data achieved an MAE of 6.314, similar to its performance in the site-target dependence scenario (Table \ref{table:harmonization_comparison}).

In this scenario, the average performance was similar across all harmonization schemes, including the \emph{No Target} scheme (Table \ref{table:harmonization_comparison}).
This result suggests that the improvement in the signal-to-noise ratio made by removing of the EoS signal was not sufficient to boost ML performance.
Consistent with our hypothesis, the \emph{No Target} scheme did not remove biologically relevant information, as this variance was shared across all sites, and the ML model showed a comparable performance as the other benchmarked schemes.

\begin{table}[h]
\centering
\caption{Comparison of performance metrics across different harmonization schemes.}
\label{table:harmonization_comparison}
\resizebox{1\textwidth}{!}{%

\begin{tabular}{|l|c|c|c|c|c|c|}
\hline
\textbf{Harmonization Scheme} & \multicolumn{3}{c|}{\textbf{Site-target dependence}} & \multicolumn{3}{c|}{\textbf{Site-target independence}} \\ \cline{2-7}
                              & \textbf{MAE} & \textbf{R\textsuperscript{2}} & \textbf{Age Bias} & \textbf{MAE} & \textbf{R\textsuperscript{2}} & \textbf{Age Bias} \\ \hline
\textbf{Unharmonized}         & 6.20         & 0.81                          & -0.43             & 6.314        & 0.785                          & -0.341            \\ \hline
\textbf{PrettYharmonize}      & 4.12         & 0.919                         & -0.26             & 6.306        & 0.769                          & -0.423            \\ \hline
\textbf{WDH}                  & 3.82         & 0.925                         & -0.32             & 6.034        & 0.803                          & -0.366            \\ \hline
\textbf{TTL}                  & 4.28         & 0.912                         & -0.23             & 6.153        & 0.775                          & -0.319            \\ \hline
\textbf{No Target}            & 15.93        & -0.007                        & -0.998            & 6.036        & 0.790                          & -0.361            \\ \hline
\end{tabular}}
\end{table}

\subsubsection{Sex classification}
In this experiment, two datasets were utilized: eNKI and CamCAN, both containing healthy controls. To enforce site-target dependence, 95\% of the females were retained from the eNKI dataset, while only 5\% were retained from the CamCAN dataset. The age ranges of the individuals were completely overlapping across the two datasets. The \emph{Unharmonized} scheme achieved a high performance (AUC = 0.97), consistent with results reported in the literature \cite{flint2020biological}.

The harmonization schemes that allow data leakage (\emph{WDH} and \emph{TTL}) and PrettYharmonize did not show any improvement over the \emph{Unharmonized} scheme (Table \ref{table:classification_performance}).
This is likely due to the presence of a strong sex-related signal in the features, which enables high performance (AUC = 0.97) even without harmonization.
Consistent with the findings from the age regression experiment, the \emph{No Target} scheme, removed sex-related information, resulting in a significant drop in classification performance (Table \ref{table:classification_performance}).

Using the same dataset generated for the age regression problem under site-target independence scenario, a sex classification experiment was conducted.
The \emph{Unharmonized} scheme achieved a slightly lower performance (AUC = 0.918) compared to the site-target dependence scenario (Table \ref{table:classification_performance}).
None of the harmonization schemes demonstrated improved classification performance relative to the \emph{Unharmonized} model (Table \ref{table:classification_performance}).
Again consistent with the age regression experiment, the \emph{No Target} scheme did not eliminate target-related variance during harmonization, leading to performance similar to the other schemes.

\begin{table}[htbp]
\centering
\caption{Comparison of sex classification performance metrics across different harmonization schemes.}
\resizebox{1\textwidth}{!}{%

\begin{tabular}{|l|c|c|c|c|c|c|}
\hline
\textbf{Harmonization Scheme} & \multicolumn{3}{c|}{\textbf{Site-target dependence}} & \multicolumn{3}{c|}{\textbf{Site-target independence}} \\ \cline{2-7}
                              & \textbf{AUC} & \textbf{bACC [\%]} & \textbf{F1} & \textbf{AUC} & \textbf{bACC [\%]} & \textbf{F1} \\ \hline
\textbf{Unharmonized}         & 0.969        & 92.64              & 0.923       & 0.918        & 84.94              & 0.851       \\ \hline
\textbf{PrettYharmonize}      & 0.968        & 92.18              & 0.918       & 0.921        & 85.06              & 0.851       \\ \hline
\textbf{WDH}                  & 0.975        & 92.10              & 0.917       & 0.913        & 84.64              & 0.847       \\ \hline
\textbf{TTL}                  & 0.967        & 92.07              & 0.917       & 0.918        & 85.16              & 0.852       \\ \hline
\textbf{No Target}            & 0.703        & 63.08              & 0.608       & 0.919        & 84.85              & 0.849       \\ \hline
\end{tabular}}
\label{table:classification_performance}
\end{table}

\subsubsection{Dementia and mild cognitive impairment classification}

For the site-target dependence scenario, 100 dementia-MCI patients and 10 controls were selected from one site, while 100 controls and 10 dementia-MCI patients were selected from a second site within the ADNI dataset. The \emph{Unharmonized} method achieved an AUC of 0.81, consistent with findings reported in the literature \cite{illakiya2023automatic}. \emph{PrettYharmonize}, \emph{WDH}, and \emph{TTL} showed a slight improvement in classification performance compared to the \emph{Unharmonized} method (Table \ref{tab:demencia_mci_performance}. As observed in previous site-target dependence experiments, the \emph{No Target} scheme removed biologically relevant information, significantly impairing the ML model's performance (Table \ref{tab:demencia_mci_performance}).

For the site-target independence scenario, an equal number of controls and patients were selected in each sites. In this case, all benchmarked schemes achieved similar classification performance across all metrics (Table \ref{tab:demencia_mci_performance}). Notably, a consistent performance drop was observed across all schemes compared to the site-target dependence scenario.

\begin{table}[h!]
\centering
\caption{Classification performance metrics for dementia-MCI prediction task in site-target dependent and independent scenarios.}
\label{tab:demencia_mci_performance}
\resizebox{1\textwidth}{!}{%

\begin{tabular}{|l|c|c|c|c|c|c|}
\hline
\textbf{Harmonization Scheme} & \multicolumn{3}{c|}{\textbf{Site-target dependence}} & \multicolumn{3}{c|}{\textbf{Site-target independence}} \\ \cline{2-7}
                              & \textbf{AUC} & \textbf{bACC [\%]} & \textbf{F1} & \textbf{AUC} & \textbf{bACC [\%]} & \textbf{F1} \\ \hline
\textbf{Unharmonized}         & 0.8131       & 73.7273            & 0.7371      & 0.7092       & 65.68              & 0.6698      \\ \hline
\textbf{PrettYharmonize}      & 0.8429       & 77.2727            & 0.7715      & 0.7089       & 65.31              & 0.6659      \\ \hline
\textbf{WDH}                  & 0.8385       & 76.6364            & 0.7644      & 0.7118       & 66.01              & 0.6755      \\ \hline
\textbf{TTL}                  & 0.8381       & 76.3636            & 0.7622      & 0.7103       & 65.85              & 0.6742      \\ \hline
\textbf{No Target}            & 0.6384       & 60.1818            & 0.6054      & 0.7096       & 66.23              & 0.6794      \\ \hline
\end{tabular}}
\end{table}

\subsubsection{Discharge status prediction of septic patients}

From the eICU dataset, 20 sites with more than 50 patients were selected. To enforce a site-target relationship, ``Alive" patients were removed from 10 sites, and ``Expired" patients were removed from the other 10 sites. The \emph{Unharmonized} scheme achieved an AUC of 0.76, slightly lower than values previously reported in the literature \cite{wernly2021machine}. This difference is expected, as fewer patients were used in our experiments compared to the referenced study.

\emph{PrettYharmonize} demonstrated an improvement in AUC performance compared to all benchmarked schemes, achieving an AUC of 0.86. In contrast, and consistent with the previous site-target dependence scenarios, the \emph{No Target} scheme removed nearly all relevant information, resulting in performance close to chance (AUC = 0.57) (Table \ref{tab:discharge_status_performance}).

For the site-target independence scenario, an equal number of ``Alive" and ``Expired" patients were retained across all 20 sites. All methods achieved similar classification performance across all metrics. The \emph{Unharmonized} method obtained a slightly lower AUC (0.72) compared to the site-target dependence scenario (0.76) (Table \ref{tab:discharge_status_performance}).

\emph{PrettYharmonize} and the leakage-prone schemes (\emph{WDH} and \emph{TTL}) showed a drop in classification performance compared to the site-target dependence scenario. The \emph{No Target} method, however, did not remove biologically relevant information during harmonization, achieving performance similar to the other benchmarked methods (AUC = 0.72).

\begin{table}[!ht]
\centering
\caption{Classification performance metrics for discharge status prediction task in site-target dependent and independent scenarios.}
\label{tab:discharge_status_performance}
\resizebox{1\textwidth}{!}{%

\begin{tabular}{|l|c|c|c|c|c|c|}
\hline
\textbf{Harmonization Scheme} & \multicolumn{3}{c|}{\textbf{Site-target dependence}} & \multicolumn{3}{c|}{\textbf{Site-target independence}} \\ \cline{2-7}
                              & \textbf{AUC} & \textbf{bACC [\%]} & \textbf{F1} & \textbf{AUC} & \textbf{bACC [\%]} & \textbf{F1} \\ \hline
\textbf{Unharmonized}         & 0.7655       & 64.37              & 0.4571      & 0.7227       & 66.88              & 0.6250      \\ \hline
\textbf{PrettYharmonize}      & 0.8588       & 66.25              & 0.4910      & 0.7101       & 66.14              & 0.6295      \\ \hline
\textbf{WDH}                  & 0.7995       & 63.39              & 0.4408      & 0.7029       & 65.25              & 0.6133      \\ \hline
\textbf{TTL}                  & 0.7897       & 63.91              & 0.4517      & 0.6907       & 64.75              & 0.6091      \\ \hline
\textbf{No Target}            & 0.5723       & 51.66              & 0.0921      & 0.7198       & 66.42              & 0.6211      \\ \hline
\end{tabular}}
\end{table}

\section*{Discussion}

In many clinical domains, data collected from a single location is insufficient for training generalizable ML models, as large and diverse datasets are needed to identify meaningful feature-target relationships. 
Combining data from multiple acquisition sites is thus a common and attractive strategy for developing ML models. 
However, this introduces unwanted variability due to acquisition idiosyncrasies. 
Data harmonization methods, such as ComBat, are often used to remove site-related variability with promising results in ML applications \cite{orlhac2022guide}. 
Integrating harmonization methods in ML pipelines brings additional challenges.
For instance, care must be taken to avoid data leakage by properly separating training and test sets \cite{kapoor2023leakage, sasse2023leakage, marzi2024efficacy}. 
Other sources of leakage and the effectiveness and practicality of data harmonization in ML workflows remain under-investigated, particularly in scenarios where the target variable and acquisition site are not independent-a situation in which harmonization methods typically struggle. 
In this study, we evaluated several harmonization schemes using biomedical datasets from various domains, specifically analyzing their performance in cases of both site-target dependence and independence.

In site-target dependence scenarios, we observed performance improvements when applying ComBat with the target specified as a covariate, which preserves its variance (Table \ref{table:harmonization_comparison},\ref{tab:demencia_mci_performance},\ref{tab:discharge_status_performance}.
However, using the target as a covariate inevitably leads to data leakage as the target values of the test set are used in both Whole Data Harmonization (\emph{WDH}) and Test Target Leakage (\emph{TTL}) schemes.
In the \emph{WDH} scheme, all available data is used to train the harmonization model, and the whole dataset is transformed before splitting it into train and test sets, leading to a form of leakage known as ``preprocessing on training and test sets" \cite{kapoor2023leakage}.
Although in the TTL scheme the test data is not used in the harmonization training process, the ComBat model still requires the target values of the test set to correctly transform the test data, a form of ``target leakage" \cite{sasse2023leakage}.
As expected, using ComBat without the target as a covariate (\emph{No Target}) avoids data leakage but removes biologically relevant information, as evidenced by the marked and consistent drop in the performance across all evaluated scenarios (Tables \ref{table:harmonization_comparison}, \ref{table:classification_performance}, \ref{tab:demencia_mci_performance}, \ref{tab:discharge_status_performance}).
These results empirically demonstrate the impact of the site-target dependence, which violates ComBat's assumption that relevant variance, i.e. target variance, is shared across sites.

Conversely, in site-target independence scenarios, despite testing a wide range of tasks and datasets from different domains, none of the harmonization schemes improved performance over the baseline approach of simply pooling the data. 
This also includes our proposed PrettYharmonize method which, while not being a harmonization method itself, uses a novel way to leverage harmonization without causing data leakage.
This lack of performance improvement may be due to the fact that the harmonization process did not sufficiently enhance the signal-to-noise ratio to allow the models to generate better predictions.
Nevertheless, we acknowledge that it is possible that our selection of data and tasks, while comprehensive, may not encompass all scenarios where harmonization could prove beneficial.

The ML pipelines pooling data without any harmonization showed performance consistent with values reported in the literature for all evaluated tasks \cite{more2023brain,illakiya2023automatic,wernly2021machine,flint2020biological}, indicating correct application of ML models.
As expected, all models using pooled unharmonized data performed better in site-target dependence scenarios, as the ML models could exploit the EoS signal, which in this case is related to the target, illicitly increasing their classification performance.

An alternative approach to training harmonization models is to use phantoms or traveling subjects \cite{ibrahim2021radiomics, maikusa2021comparison}.
This can allow for accurate estimation of location and scale parameters specific to each acquisition setup and parameter setting, which can then be applied to real data.
This approach mitigates the risk of inadvertently removing meaningful biological variation during harmonization.
However, it is domain-specific and needs additional data collection which is costly as well as accounting for other challenges such as temporal shifts in equipment.

With respect to clinical data, to our knowledge, this is the first study to explore the potential of applying harmonization to this type of data.
It is important to distinguish this form of harmonization, which aims to remove EoS from features, from the more common use of the term in the clinical domain, where ``harmonization" typically refers to the standardization of feature names and units across sites \cite{plebani2016harmonization}.

The proposed \emph{PrettYharmonize} method is designed to avoid data leakage.
It first harmonizes the data using pretended labels and uses the harmonized output to learn a prediction model.
The main idea behind \emph{PrettYharmonize} is learning subtle differences caused by use of correct or incorrect labels.
This design ensures that the harmonization process remains confined to the training phase and circumvents the need for actual target values from the test samples, ensuring a leakage-free ML pipeline.
Several key aspects of \emph{PrettYharmonize} deserve emphasis.
First, it is built around the neuroHarmonize method, meaning that we do not propose a novel harmonization method but rather introduce an approach that internally integrates harmonization into ML pipelines while avoiding data leakage.
This distinction is crucial, as the innovation lies not in altering the harmonization process but in structuring its application to ensure compatibility with ML workflows across both site-target dependence scenarios.
Furthermore, this modular design of \emph{PrettYharmonize} enables the use of other harmonization methods.
Second, the output of \emph{PrettYharmonize} is not a harmonized dataset but rather a final target prediction.
Unlike traditional harmonization schemes, which produce a harmonized dataset for downstream use, our method uses harmonized data internally as an intermediate step to generate a prediction.
By focusing on prediction, the method prioritizes practical utility in real-world applications where the primary goal is accurate target prediction rather than data transformation.

\emph{PrettYharmonize} was rigorously validated on the MAREoS datasets, which were specifically designed to evaluate harmonization methods \cite{solanes2023mareos}.
The strong performance observed on these benchmark datasets indicates that the internal harmonization scheme, based on pretended targets, successfully harmonizes data for the posterior use of the predictive model.
In addition, on real-world datasets \emph{PrettYharmonize} achieved competitive performance comparable to leakage-prone pipelines, demonstrating its practical utility.
These results suggest that the proposed method is a promising alternative for real-world applications, particularly in settings where data leakage poses a substantial risk.
Consequently, we advocate for its adoption in future use cases. 
Taken together, our findings underscore the need for meticulous evaluation when using data harmonization together with machine learning and encourage adoption of reproducible, open science practices to advance the field and benefit the wider community.

\subsection*{Conclusion}

This study highlights three key findings regarding the use of harmonization methods in ML pipelines.
First, while ComBat-based methods do not intrinsically cause data leakage, they might require that target variance be preserved as a covariate during training when a site-target relationship exists.
As demonstrated, this requirement precludes their use in real-world ML applications, as it requires target values of the test samples that are not available.
Our results empirically demonstrate that failing to preserve target variance leads to the removal of relevant signals, significantly undermining the effectiveness of harmonization.
Second, harmonization did not lead to meaningful performance improvement when the site and targets are independent.
This suggests that the benefits of harmonization may be limited in such cases, as the removal of site-related variability might not sufficiently enhance the signal-to-noise ratio for ML models.
Finally, the proposed PrettYharmonize method advances the field by adapting neuroHarmonize for ML pipelines, effectively integrating harmonization while eliminating the need for test targets during training or transformation.
This approach not only prevents data leakage by design but also achieved encouraging results, ensuring that the harmonization process remains both robust and practical for real-world applications.

\subsection*{Limitations of the study}

While this study provides valuable insights into harmonization methods, several limitations must be acknowledged.
First, although our analysis focused on widely used ComBat-based techniques, other harmonization approaches, such as deep learning-based methods (e.g., style-matching generative models or variational autoencoders \cite{hu2023review, abbasi2024deep}), were not explored.
These methods may offer promising alternatives, particularly in complex scenarios where traditional approaches fall short, although they often require significant computational resources and large datasets.

Second, the impact of harmonization on feature selection and model interpretability has not been thoroughly investigated.
Future research should explore how harmonization methods affect model behavior, especially in different domains and contexts, to better understand their broader implications for ML pipelines.

Third, while we simulated extreme site-target dependence and independence scenarios in a controlled manner, real-world cases are likely to fall somewhere along this spectrum.
Our goal was to highlight potential problems that may arise from applying harmonization without proper consideration.
Further studies are needed to investigate the effectiveness of harmonization methods at varying degrees of site-target dependence.

Finally, the proposed method introduces additional computational complexity compared to training a single neuroHarmonize model, as the ``pretending" process requires multiple harmonization steps.
However, the extra burden of this process is limited to \emph{applying} a trained neuroHarmonize model several times (one harmonization transformation for each presented class), which is less time-consuming than training the model. 
Importantly, the computational cost of the pretending process does not scale with the number of sites to be harmonized, mitigating some of the practical limitations.

\section*{Experimental procedures}

\subsubsection*{Data and code availability}

All used MRI datasets are publicly available, possibly upon registration. 
The eICU dataset is publicly available at \url{https://physionet.org/content/eicu-crd/2.0/} after registration. 
Registration includes the completion of a training course in research with human individuals at \url{https://about.citiprogram.org/} and signing of a data use agreement mandating responsible handling of the data and adhering to the principle of collaborative research.

The \emph{PrettYharmonize} library is publicly available at: \url{https://github.com/juaml/PrettYharmonize}. 
The scripts to replicate the experiments and to process the datasets are available at: \url{https://github.com/juaml/harmonize_project}

\subsection*{Data description}
\subsection*{MAREoS dataset}
To ensure the validity of PrettYharmonize, we benchmarked it in a classification problem using the datasets specifically designed to evaluate harmonization models \cite{solanes2023mareos}. This MAREoS dataset consists of eight datasets simulating 18 MRI features (cortical thickness, cortical surface area, or subcortical volumes). Four datasets contain a “True” signal and four only contain an Effect of Site (“EoS”) signal related to a binary target. In that sense, an ML model that learns the “EoS” signal can fraudulently achieve good classification performance. The signal, both the True or EoS, are called “Simple” and “Interactions”, depending on a linear or non-linear effect, respectively. Within each dataset, approximately 1000 samples, coming from 8 sites, were simulated. The datasets are provided as 10 train and test fold pairs. For the dataset containing only the EoS, the methods should be able to remove this effect and the classification performance should be at the chance level, i.e. balanced accuracy (bACC) of 50\%. On the other hand, in the dataset with only the True signal, the harmonization models should not degrade the signal, and the bACC is expected to be a high value (bACC $\approx$ 80\%).

\subsection*{MRI data}

To empirically compare different harmonization schemes with and without site-target dependence, age regression, and sex classification were performed using MRI data. These targets were used as they are highly reliable and can be easily obtained. For all T1-weighted MR images, Voxel-Based Morphometry was performed using CAT12.8 \cite{gaser2022cat} to obtain modulated gray matter (GM) volume, which was then linearly resampled to 8x8x8 mm3 voxels, resulting in 3747 voxels that were used as features. Five datasets were used: Amsterdam Open MRI Collection (AOMIC-ID1000) \cite{snoek2021amsterdam}, The Enhanced Nathan Kline Institute (eNKI) \cite{nooner2012nki}, Cambridge Centre for Ageing Neuroscience (CamCAN) \cite{shafto2014cambridge}, 1000Brains \cite{caspers2014studying}, and the Southwest University Adult Lifespan Dataset (SALD) \cite{wei2018structural}. These datasets were selected as the data within each dataset was acquired only in one site thus avoiding additional confounding. The demographic information of these datasets is presented in Table \ref{table:datasets}.

\begin{table}[htbp]
\centering
\caption{Characteristics of the original MRI datasets used in the study.}
\resizebox{1\textwidth}{!}{%

\begin{tabular}{|l|c|c|c|c|c|c|}
\hline
\textbf{Dataset Name} & \textbf{N Images} & \textbf{Mean Age} & \textbf{Std Age} & \textbf{Min Age} & \textbf{Max Age} & \textbf{\% Female} \\ \hline
AOMIC-ID1000          & 928 & 22.85 & 1.71 & 19 & 26 & 52\% \\\hline
eNKI                  & 818 & 46.90 & 17.73 & 19 & 85              & 65\%\\ \hline
CamCAN                & 651 & 54.27 & 18.59 & 18 & 88              & 50\%\\ \hline
1000Brains            & 1144 & 61.84 & 12.39 & 21 & 85              & 55\%\\ \hline
SALD                  & 494 & 45.18 & 17.44 & 19 & 80              & 62\%\\ \hline
\end{tabular}}
\label{table:datasets}
\end{table}

\subsubsection*{Age regression }

\paragraph{Forced site-target dependence.}

Four datasets, AOMIC-ID1000, eNKI, CamCAN, and 1000Brains, were randomly subsampled in different age ranges, forcing a site-target dependence. The subsample was performed to ensure the same amount of subjects by each sex in each dataset (Table \ref{table:site_target_dependent}). 

\begin{table}[htbp]
\centering
\caption{Dataset characteristics for site-target dependent age regression experiment.}
\resizebox{1\textwidth}{!}{%

\begin{tabular}{|l|c|c|c|c|c|c|}
\hline
\textbf{Dataset Name} & \textbf{N Images} & \textbf{Mean Age} & \textbf{Std Age} & \textbf{Min Age} & \textbf{Max Age} & \textbf{\% Female} \\ \hline
AOMIC-ID1000          & 118               & 22.73             & 1.61             & 19              & 26              & 50\%              \\ \hline
eNKI                  & 118               & 33.00             & 3.95             & 27              & 40              & 50\%              \\ \hline
CamCAN                & 118               & 50.09             & 6.06             & 41              & 60              & 50\%              \\ \hline
1000Brains            & 118               & 68.74             & 5.02             & 61              & 79              & 50\%              \\ \hline
\end{tabular}}
\label{table:site_target_dependent}
\end{table}

\paragraph{Forced site-target independence.}

Three datasets, CamCAN, eNIKI, and SALD, were used. The datasets were selected as they contain individuals covering a wide range of ages above 18. The AOMIC and 1000Brains datasets were excluded as those datasets mainly included young and old participants, respectively. Each dataset was balanced in terms of sex and age (Table \ref{table:independent_experiments}). This was achieved by retaining the same number of subjects for each sex in 10 equally distributed age ranges, from the minimum to the maximum age in each dataset.

\begin{table}[htbp]
\centering

\caption{Dataset characteristics for site-target independent age regression and sex classification experiments.}
\resizebox{1\textwidth}{!}{%

\begin{tabular}{|l|c|c|c|c|c|c|}
\hline
\textbf{Dataset} & \textbf{N Images} & \textbf{Mean Age} & \textbf{Std Age} & \textbf{Min Age} & \textbf{Max Age} & \textbf{\% Female} \\ \hline
SALD             & 200               & 48.99             & 16.97            & 19              & 77              & 50\%              \\ \hline
eNKI             & 300               & 47.75             & 17.43            & 18              & 78              & 50\%              \\ \hline
CamCAN           & 288               & 48.60             & 18.00            & 18              & 78              & 50\%              \\ \hline
\end{tabular}}
\label{table:independent_experiments}
\end{table}

\subsubsection{Sex classification}

For this experiment, only the eNKI and CamCAN datasets were used, as those present a broad and similar age range. In this case, the percentages of females in each dataset were forced to be 95\% in eNKI and 5\% in CamCAN (Table \ref{table:sex_classification}). Additionally, the same number of images for each dataset was retained.

\begin{table}[htbp]
\centering

\caption{Dataset characteristics for site-target dependent sex classification experiment.}
\resizebox{1\textwidth}{!}{%

\begin{tabular}{|l|c|c|c|c|c|c|}
\hline
\textbf{Dataset Name} & \textbf{N Images} & \textbf{Mean Age} & \textbf{Std Age} & \textbf{Min Age} & \textbf{Max Age} & \textbf{\% Female} \\ \hline
eNKI                  & 295               & 45.13             & 18.93            & 19              & 84              & 5\%               \\ \hline
CamCAN                & 295               & 53.77             & 18.51            & 18              & 88              & 95\%              \\ \hline
\end{tabular}}
\label{table:sex_classification}
\end{table}

\paragraph{Forced site-target independence}

The same dataset generated in the site-target independence scenario for age regression was used for sex classification.

\subsubsection{Dementia and mild cognitive impairment classification}
\paragraph{Forced site-target dependence}

Data used in the preparation of this article were obtained from the Alzheimer’s Disease Neuroimaging Initiative (ADNI) database (adni.loni.usc.edu). The ADNI was launched in 2003 as a public-private partnership, led by Principal Investigator Michael W. Weiner, MD. The primary goal of ADNI has been to test whether serial magnetic resonance imaging (MRI), positron emission tomography (PET), other biological markers, and clinical and neuropsychological assessment can be combined to measure the progression of mild cognitive impairment (MCI) and early Alzheimer’s disease (AD).

In our experiments using the ADNI dataset, where the data were acquired in different sites, we selected 100 dementia-MCI patients and 10 controls from one site. We selected 100 control patients and 10 dementia-MCi patients from another site, again forcing the site-target relationship (Table \ref{tab:d-mci_dependent}). The images were processed with FreeSurfer \cite{fischl2012freesurfer}. The thickness of 74 cerebral and sub-cerebral structures were extracted as features.

\begin{table}
\centering
\caption{Datasets characteristics for site-target dependent Dementia-MCI classification experiment}
\label{tab:d-mci_dependent}
\resizebox{1\textwidth}{!}{%

\begin{tabular}{|c|c|c|c|c|c|c|}
\hline
\textbf{Dataset name} & \textbf{N Images} & \textbf{Mean Age} & \textbf{Std Age} & \textbf{Min Age} & \textbf{Max Age} & \textbf{\% Dementia-MCI} \\ \hline
\textbf{Site 1}       & 110               & 75.182            & 6.667            & 59               & 92               & 9 \%                      \\ \hline
\textbf{Site 2}       & 110               & 72.331            & 5.560            & 60               & 97               & 91 \%                     \\ \hline
\end{tabular}
}
\end{table}

\paragraph{Forced site-target independence}

Using the ADNI dataset, the same extracted features were used. From the dataset, 126 dementia-MCI and control patients were randomly selected from the first site while 56  dementia-MCI and control patients were randomly selected from the second site (Table \ref{tab:independent_dementia_mci_dataset_characteristics}).  

\begin{table}[htbp]
\caption{Datasets characteristics for site-target independent dementia-MCI classification experiment}
\label{tab:independent_dementia_mci_dataset_characteristics}
\centering
\resizebox{1\textwidth}{!}{%

\begin{tabular}{|c|c|c|c|c|c|c|}
\hline
\textbf{Dataset ID} & \textbf{N Images} & \textbf{Mean Age} & \textbf{Std Age} & \textbf{Min Age} & \textbf{Max Age} & \textbf{\% Dementia-MCI} \\ \hline
\textbf{Site 1}     & 252               & 73.72             & 6.582            & 59               & 93               & 50 \%                     \\ \hline
\textbf{Site 2}     & 114               & 72.68             & 6.448            & 56               & 96               & 50 \%                     \\ \hline
\end{tabular}
}

\end{table}

\subsubsection{Discharge status prediction of septic patients}
\subsubsection{Forced site-target dependence}

The eICU \cite{pollard2019eicu,pollard2018eicu,goldberger2000physiobank}  dataset was used for the experiments, which contains 200859 ICU stays from 139367 patients in 208 different ICUs across the United States. We use a well-known problem of classified hospital discharge (Expired or Alive), in septic patients \cite{hou2020predicting,deng2022evaluating}. The approach described in \cite{wernly2021machine} was followed for selecting the features and extracting the patient cohort. The features used were arterial blood gases: paO2, paCO2, pH, base excess, Hgb, glucose, bicarbonate, and lactate. After the patients’ selection, a final dataset of 496 Expired and 3021 Alive patients was retained. From this filtered dataset, we remove those sites with less than 50 stays, retaining 20 final sites. 

From 20 of these sites, all “Alive” patients, except for one, were removed for the 10 sites with more “Expired” patients. On the contrary, all the “Expired” patients, except for one, were removed from the 10 sites with fewer “Expired” patients. Note that in this case, the classes (Alive and Expired) are represented in several sites and are not the same in each site (Table \ref{tab:outcome_prediction_dataset_characteristics_dependent}), compared with the previous classification experiments performed. A total of 249 Expired and 666 Alive patients were used in this experiment.

\begin{table}
\centering
\caption{Datasets characteristics for site-target dependent outcome prediction on septic patients experiment}
\label{tab:outcome_prediction_dataset_characteristics_dependent}
\resizebox{1\textwidth}{!}{%

\begin{tabular}{|c|c|c|c|c|}
\hline
\textbf{Dataset ID} & \textbf{N Images} & \textbf{Alive Count} & \textbf{Expired Count} & \textbf{\% Expired} \\ \hline
\textbf{Site 79}    & 92                & 91                   & 1                      & 1.08 \%             \\ \hline
\textbf{Site 148}   & 69                & 68                   & 1                      & 1.14 \%             \\ \hline
\textbf{Site 15}    & 56                & 55                   & 1                      & 1.17 \%             \\ \hline
\textbf{Site 157}   & 14                & 1                    & 13                     & 92.85 \%            \\ \hline
\textbf{Site 165}   & 15                & 1                    & 14                     & 93.33 \%            \\ \hline
\textbf{Site 167}   & 21                & 1                    & 20                     & 95.54 \%            \\ \hline
\textbf{Site 176}   & 111               & 110                  & 1                      & 0.90 \%             \\ \hline
\textbf{Site 188}   & 32                & 1                    & 31                     & 96.87 \%            \\ \hline
\textbf{Site 248}   & 55                & 54                   & 1                      & 1.82 \%             \\ \hline
\textbf{Site 252}   & 27                & 1                    & 26                     & 96.30 \%            \\ \hline
\textbf{Site 264}   & 17                & 1                    & 16                     & 94.12 \%            \\ \hline
\textbf{Site 300}   & 69                & 68                   & 1                      & 1.45 \%             \\ \hline
\textbf{Site 345}   & 55                & 54                   & 1                      & 1.82 \%             \\ \hline
\textbf{Site 365}   & 58                & 57                   & 1                      & 1.72 \%             \\ \hline
\textbf{Site 416}   & 15                & 1                    & 14                     & 93.33 \%            \\ \hline
\textbf{Site 420}   & 65                & 1                    & 64                     & 98.46 \%            \\ \hline
\textbf{Site 443}   & 58                & 57                   & 1                      & 1.72 \%             \\ \hline
\textbf{Site 449}   & 18                & 1                    & 17                     & 94.44 \%            \\ \hline
\textbf{Site 452}   & 43                & 42                   & 1                      & 2.33 \%             \\ \hline
\textbf{Site 458}   & 24                & 1                    & 23                     & 95.83 \%            \\ \hline
\textbf{Total}      & 915               & 666                  & 249                    & 27.21 \%            \\ \hline
\end{tabular}}
\end{table}

\subsubsection{Forced site-target independence}

The same feature extraction and patient selection were made (496 Expired and 3021 Alive patients) from the eICU. The same 20 sites, with more than 50 images, were used.

From all sites, the same number of “Alive” and “Expired” patients were retained. A total of 324 Expired and 324 Alive patients were used in this experiment (Table \ref{tab:independent_outcome_prediction_dataset_characteristics}).

\begin{table}
\centering
\caption{Datasets characteristics for site-target independent outcome prediction on septic patients experiment}
\label{tab:independent_outcome_prediction_dataset_characteristics}
\resizebox{1\textwidth}{!}{%

\begin{tabular}{|c|c|c|c|c|}
\hline
\textbf{Dataset ID} & \textbf{N Images} & \textbf{Alive Count} & \textbf{Expired Count} & \textbf{\% Expired} \\ \hline
\textbf{Site 79}    & 20                & 10                   & 10                     & 50 \%               \\ \hline
\textbf{Site 148}   & 24                & 12                   & 12                     & 50 \%               \\ \hline
\textbf{Site 154}   & 18                & 9                    & 9                      & 50 \%               \\ \hline
\textbf{Site 157}   & 26                & 13                   & 13                     & 50 \%               \\ \hline
\textbf{Site 165}   & 28                & 14                   & 14                     & 50 \%               \\ \hline
\textbf{Site 167}   & 42                & 21                   & 21                     & 50 \%               \\ \hline
\textbf{Site 176}   & 14                & 7                    & 7                      & 50 \%               \\ \hline
\textbf{Site 188}   & 62                & 31                   & 31                     & 50 \%               \\ \hline
\textbf{Site 248}   & 26                & 13                   & 13                     & 50 \%               \\ \hline
\textbf{Site 252}   & 52                & 26                   & 26                     & 50 \%               \\ \hline
\textbf{Site 264}   & 32                & 16                   & 16                     & 50 \%               \\ \hline
\textbf{Site 300}   & 12                & 6                    & 6                      & 50 \%               \\ \hline
\textbf{Site 345}   & 8                 & 4                    & 4                      & 50 \%               \\ \hline
\textbf{Site 365}   & 8                 & 4                    & 4                      & 50 \%               \\ \hline
\textbf{Site 416}   & 28                & 14                   & 14                     & 50 \%               \\ \hline
\textbf{Site 420}   & 128               & 64                   & 64                     & 50 \%               \\ \hline
\textbf{Site 443}   & 22                & 11                   & 11                     & 50 \%               \\ \hline
\textbf{Site 449}   & 34                & 17                   & 17                     & 50 \%               \\ \hline
\textbf{Site 452}   & 18                & 9                    & 9                      & 50 \%               \\ \hline
\textbf{Site 458}   & 46                & 23                   & 23                     & 50 \%               \\ \hline
\textbf{Total}      & 648               & 324                  & 324                    & 50 \%               \\ \hline
\end{tabular}}
\end{table}

\subsection*{Methods}

\subsection*{PrettYHarmonize}

\begin{figure}
  \centering
    \includegraphics[width=0.70\linewidth]{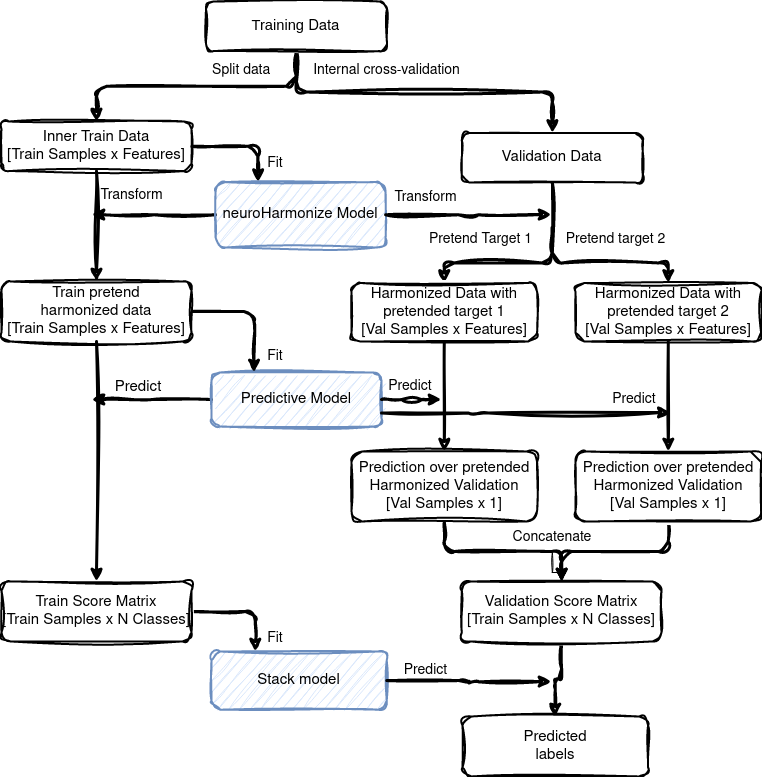}
\caption{PrettYharmonize training workflow. The workflow showcases the training workflow for a binary classification problem.}
  \label{fig:workflow}
\end{figure}

The proposed method introduces a new harmonization scheme based on the neuroHarmonize model. This scheme relies on the use of ``pretended" target values during the harmonization process, enabling predictions to be made without requiring actual test target values (Figure \ref{fig:workflow}). For our experiments, the available data is split into training and test sets, simulating a real-world use case. The training fold is further divided into inner training and validation folds. A neuroHarmonize model is trained on the inner training data to learn how to remove the site’s effect. The inner training data is then harmonized, and a predictive model is trained on the harmonized inner training data to predict the target. This predictive model must be selected as the best possible model to address the specific problem at hand.

Using the trained neuroHarmonize model, the validation samples are harmonized while “pretending” their target value. For example, for a binary classification problem, all the validation labels are set as the first class, pretending that all validation samples belong to the first class. Using these “pretended” labels, the validation data is harmonized and the predictions are made using the trained Predictive model. Later, the validation labels are set to the second class and the validation data is harmonized again and a new prediction is generated. In general, for a classification task, the set of available classes is pretended, while for a regression task, the values are linearly sampled in the target’s range. All the predictions, generated with the pretended harmonized data, are concatenated and a “Score matrix” is created. This matrix has a dimension of number of validation samples times the number of labels. To effectively utilize the training dataset, a K-fold cross-validation (CV) procedure was employed, generating out-of-sample predictions for the entire dataset. Using this Score matrix as input features, a “Stack” model is trained to predict the target and give a final prediction. 

At the test time, when the test label is not available, the same procedure is followed.  For example, in a binary classification problem, a test sample will be harmonized using the neuroHarmonize model first pretending that the test label belongs to the first class. The Predict model will generate a prediction using the harmonized data, and the process will be repeated pretending the test sample belongs to the second class. Both predictions generated by the Predictive model are concatenated and a test Score matrix is built. This matrix is used by the Stack model, which generates the final prediction.

\clearpage
\subsection{Machine learning model}

For the binary classification problem using the synthetic data (MAREoS datasets), a Random Forest model (RF) \cite{breiman2001random}, with default sklearn parameters, was used as a Predictive Model, and a Logistic Regression (LG) \cite{tolles2016logistic} was used as a Stack Model. The same RF model was used to train a model with the original data to obtain a classification baseline for each dataset (Baseline model).

For the age regression problems using real MRI data, Relevance Vector Regression \cite{tipping1999relevance} with a polynomial kernel of degree 1 (RVR) was used as a Predictive and Stack model for PrettYharmonize. The RVR model was used for the rest of the harmonization schemes \cite{more2023brain}. A 5-fold cross-validation scheme was used, and the Mean Absolute Error (MAE), coefficient of determination (R2), and age bias (Pearson's correlation between the true age and the difference between the predicted and true age) were calculated on the test sets. 

For the sex classification using real MRI data, RVR was used as a Predictive and Stack model. A 5 times repeated 5-fold stratified cross-validation scheme was used, and the Area under the receiver operating curve (AUC), balanced accuracy (bACC) and were calculated on the test 

\section*{Supplemental information}

In the Supplementary Information Tables S1-S3 and Figures S1-S5 and their legends are presented.

\section*{Acknowledgments}

This work was supported by the MODS project funded from the program “Profilbildung 2020” (grant no. PROFILNRW-2020-107-A), an initiative of the Ministry of Culture and Science of the State of Northrhine Westphalia.
Also by Helmholtz Portfolio Theme Supercomputing and Modeling for the Human Brain. This work was partly supported by the Helmholtz-AI project DeGen (ZT-I-PF-5-078)

Data collection and sharing for this project was funded by the Alzheimer's Disease Neuroimaging Initiative (ADNI) (National Institutes of Health Grant U01 AG024904) and DOD ADNI (Department of Defense award number W81XWH-12-2-0012). ADNI is funded by the National Institute on Aging, the National Institute of Biomedical Imaging and Bioengineering, and through generous contributions from the following: AbbVie, Alzheimer’s Association; Alzheimer’s Drug Discovery Foundation; Araclon Biotech; BioClinica, Inc.; Biogen; Bristol-Myers Squibb Company; CereSpir, Inc.; Cogstate; Eisai Inc.; Elan Pharmaceuticals, Inc.; Eli Lilly and Company; EuroImmun; F. Hoffmann-La Roche Ltd and its affiliated company Genentech, Inc.; Fujirebio; GE Healthcare; IXICO Ltd.; Janssen Alzheimer Immunotherapy Research \& Development, LLC.; Johnson \& Johnson Pharmaceutical Research \& Development LLC.; Lumosity; Lundbeck; Merck \& Co., Inc.; Meso Scale Diagnostics, LLC.; NeuroRx Research; Neurotrack Technologies; Novartis Pharmaceuticals Corporation; Pfizer Inc.; Piramal Imaging; Servier; Takeda Pharmaceutical Company; and Transition Therapeutics. The Canadian Institutes of Health Research is providing funds to support ADNI clinical sites in Canada. Private sector contributions are facilitated by the Foundation for the National Institutes of Health (www.fnih.org). The grantee organization is the Northern California Institute for Research and Education, and the study is coordinated by the Alzheimer’s Therapeutic Research Institute at the University of Southern California. ADNI data are disseminated by the Laboratory for Neuro Imaging at the University of Southern California.

\section*{Author contributions}

NN: conceptualization, data curation, formal analysis, investigation, methodology, software, validation, visualization, writing – original draft, and writing – review and editing

\noindent SE: conceptualization, formal analysis, funding acquisition, investigation, methodology, project administration, resources, supervision, writing – review and editing

\noindent CJ: conceptualization, data curation, formal analysis, funding acquisition, investigation, methodology, project administration, resources, supervision, validation, visualization, writing – review and editing

\noindent MR: funding acquisition, resources, review and editing

\noindent KD: resources, review and editing

\noindent MK: funding acquisition, resources, review and editing

\noindent AL: funding acquisition, resources, review and editing

\noindent FR: conceptualization, formal analysis, investigation, methodology, resources, software, supervision, validation, visualization, writing – original draft, and writing – review and editing

\noindent  KP: conceptualization, funding acquisition, formal analysis, investigation, methodology, project administration, supervision, validation, visualization, writing – original draft, and writing – review and editing

\section*{Declaration of interests}

The authors declare no competing interests.


\begin{thebibliography}{00}
\bibitem{hu2023review}Hu, F., Chen, A., Horng, H., Bashyam, V., Davatzikos, C., Alexander-Bloch, A., Li, M., Shou, H., Satterthwaite, T., Yu, M. \& Others Image harmonization: A review of statistical and deep learning methods for removing batch effects and evaluation metrics for effective harmonization. {\em NeuroImage}. \textbf{274} pp. 120125 (2023)
\bibitem{bayer2022site}Bayer, J., Thompson, P., Ching, C., Liu, M., Chen, A., Panzenhagen, A., Jahanshad, N., Marquand, A., Schmaal, L. \& Sämann, P. Site effects how-to and when: An overview of retrospective techniques to accommodate site effects in multi-site neuroimaging analyses. {\em Frontiers In Neurology}. \textbf{13} pp. 923988 (2022)
\bibitem{botvinik2023reproducibility}Botvinik-Nezer, R. \& Wager, T. Reproducibility in neuroimaging analysis: challenges and solutions. {\em Biological Psychiatry: Cognitive Neuroscience And Neuroimaging}. \textbf{8}, 780-788 (2023)
\bibitem{plebani2016harmonization}Plebani, M. Harmonization of clinical laboratory information–current and future strategies. {\em Ejifcc}. \textbf{27}, 15 (2016)
\bibitem{maikusa2021comparison}Maikusa, N., Zhu, Y., Uematsu, A., Yamashita, A., Saotome, K., Okada, N., Kasai, K., Okanoya, K., Yamashita, O., Tanaka, S. \& Others Comparison of traveling-subject and ComBat harmonization methods for assessing structural brain characteristics. {\em Human Brain Mapping}. \textbf{42}, 5278-5287 (2021)
\bibitem{orlhac2022guide}Orlhac, F., Eertink, J., Cottereau, A., Zijlstra, J., Thieblemont, C., Meignan, M., Boellaard, R. \& Buvat, I. A guide to ComBat harmonization of imaging biomarkers in multicenter studies. {\em Journal Of Nuclear Medicine}. \textbf{63}, 172-179 (2022)
\bibitem{abbasi2024deep}Abbasi, S., Lan, H., Choupan, J., Sheikh-Bahaei, N., Pandey, G. \& Varghese, B. Deep learning for the harmonization of structural MRI scans: a survey. {\em BioMedical Engineering OnLine}. \textbf{23}, 90 (2024)
\bibitem{hosseini2024effect}Hosseini, S., Shiri, I., Ghaffarian, P., Hajianfar, G., Avval, A., Seyfi, M., Servaes, S., Rosa-Neto, P., Zaidi, H. \& Ay, M. The effect of harmonization on the variability of PET radiomic features extracted using various segmentation methods. {\em Annals Of Nuclear Medicine}. pp. 1-15 (2024)
\bibitem{chen2014exploration}Chen, J., Liu, J., Calhoun, V., Arias-Vasquez, A., Zwiers, M., Gupta, C., Franke, B. \& Turner, J. Exploration of scanning effects in multi-site structural MRI studies. {\em Journal Of Neuroscience Methods}. \textbf{230} pp. 37-50 (2014)
\bibitem{solanes2023mareos}Solanes, A., Gosling, C., Fortea, L., Ortuño, M., Lopez-Soley, E., Llufriu, S., Madero, S., Martinez-Heras, E., Pomarol-Clotet, E., Solana, E. \& Others Removing the effects of the site in brain imaging machine-learning–Measurement and extendable benchmark. {\em NeuroImage}. \textbf{265} pp. 119800 (2023)
\bibitem{li2020denoising}Li, H., Smith, S., Gruber, S., Lukas, S., Silveri, M., Hill, K., Killgore, W. \& Nickerson, L. Denoising scanner effects from multimodal MRI data using linked independent component analysis. {\em Neuroimage}. \textbf{208} pp. 116388 (2020)
\bibitem{wachinger2021detect}Wachinger, C., Rieckmann, A., Pölsterl, S., Initiative, A. \& Others Detect and correct bias in multi-site neuroimaging datasets. {\em Medical Image Analysis}. \textbf{67} pp. 101879 (2021)
\bibitem{da2020review} Da-Ano, R., Visvikis, D. \& Hatt, M. Harmonization strategies for multicenter radiomics investigations. {\em Physics in Medicine \& Biology}. \textbf{65}, 24TR02 (2020).
\bibitem{johnson2007adjusting}Johnson, W., Li, C. \& Rabinovic, A. Adjusting batch effects in microarray expression data using empirical Bayes methods. {\em Biostatistics}. \textbf{8}, 118-127 (2007)
\bibitem{fortin2017harmonization}Fortin, J., Parker, D., Tunç, B., Watanabe, T., Elliott, M., Ruparel, K., Roalf, D., Satterthwaite, T., Gur, R., Gur, R. \& Others Harmonization of multi-site diffusion tensor imaging data. {\em Neuroimage}. \textbf{161} pp. 149-170 (2017)
\bibitem{fortin2018harmonization}Fortin, J., Cullen, N., Sheline, Y., Taylor, W., Aselcioglu, I., Cook, P., Adams, P., Cooper, C., Fava, M., McGrath, P. \& Others Harmonization of cortical thickness measurements across scanners and sites. {\em Neuroimage}. \textbf{167} pp. 104-120 (2018)
\bibitem{pomponio2019harmonization}Pomponio, R., Erus, G., Habes, M., Doshi, J., Srinivasan, D., Mamourian, E., Bashyam, V., Nasrallah, I., Satterthwaite, T., Fan, Y. \& Others Harmonization of large MRI datasets for the analysis of brain imaging patterns throughout the lifespan. NeuroImage, 208, Article 116450.  (2019)
\bibitem{leek2012sva}Leek, J., Johnson, W., Parker, H., Jaffe, A. \& Storey, J. The sva package for removing batch effects and other unwanted variation in high-throughput experiments. {\em Bioinformatics}. \textbf{28}, 882-883 (2012)
\bibitem{radua2020increased}Radua, J., Vieta, E., Shinohara, R., Kochunov, P., Quidé, Y., Green, M., Weickert, C., Weickert, T., Bruggemann, J., Kircher, T. \& Others Increased power by harmonizing structural MRI site differences with the ComBat batch adjustment method in ENIGMA. {\em Neuroimage}. \textbf{218} pp. 116956 (2020)
\bibitem{yu2018statistical}Yu, M., Linn, K., Cook, P., Phillips, M., McInnis, M., Fava, M., Trivedi, M., Weissman, M., Shinohara, R. \& Sheline, Y. Statistical harmonization corrects site effects in functional connectivity measurements from multi-site fMRI data. {\em Human Brain Mapping}. \textbf{39}, 4213-4227 (2018)
\bibitem{dudley2023abcd_harmonizer}Dudley, J., Maloney, T., Simon, J., Atluri, G., Karalunas, S., Altaye, M., Epstein, J. \& Tamm, L. ABCD\_Harmonizer: An Open-source Tool for Mapping and Controlling for Scanner Induced Variance in the Adolescent Brain Cognitive Development Study. {\em Neuroinformatics}. \textbf{21}, 323-337 (2023)
\bibitem{ibrahim2021radiomics}Ibrahim, A., Primakov, S., Beuque, M., Woodruff, H., Halilaj, I., Wu, G., Refaee, T., Granzier, R., Widaatalla, Y., Hustinx, R. \& Others Radiomics for precision medicine: Current challenges, future prospects, and the proposal of a new framework. {\em Methods}. \textbf{188} pp. 20-29 (2021)
\bibitem{acquitter2022radiomics}Acquitter, C., Piram, L., Sabatini, U., Gilhodes, J., Moyal Cohen-Jonathan, E., Ken, S. \& Lemasson, B. Radiomics-based detection of radionecrosis using harmonized multiparametric MRI. {\em Cancers}. \textbf{14}, 286 (2022)
\bibitem{barth2023vivo}Barth, C., Kelly, S., Nerland, S., Jahanshad, N., Alloza, C., Ambrogi, S., Andreassen, O., Andreou, D., Arango, C., Baeza, I. \& Others In vivo white matter microstructure in adolescents with early-onset psychosis: a multi-site mega-analysis. {\em Molecular Psychiatry}. \textbf{28}, 1159-1169 (2023)
\bibitem{bourbonne2021development}Bourbonne, V., Jaouen, V., Nguyen, T., Tissot, V., Doucet, L., Hatt, M., Visvikis, D., Pradier, O., Valéri, A., Fournier, G. \& Others Development of a radiomic-based model predicting lymph node involvement in prostate cancer patients. {\em Cancers}. \textbf{13}, 5672 (2021)
\bibitem{campello2022minimising} Campello, V., Mart\'{\i}n-Isla, C., Izquierdo, C., Guala, A., Palomares, J., Vilad\'{e}s, D., Descalzo, M., Karakas, M., \c{C}avu\c{s}, E., Raisi-Estabragh, Z. \& Others. Minimising multi-centre radiomics variability through image normalisation: a pilot study. {\em Scientific Reports}. \textbf{12}, 12532 (2022).
\bibitem{chen2023four}Chen, P., Yao, H., Tijms, B., Wang, P., Wang, D., Song, C., Yang, H., Zhang, Z., Zhao, K., Qu, Y. \& Others Four distinct subtypes of Alzheimer’s disease based on resting-state connectivity biomarkers. {\em Biological Psychiatry}. \textbf{93}, 759-769 (2023)
\bibitem{sasse2023leakage}Sasse, L., Nicolaisen-Sobesky, E., Dukart, J., Eickhoff, S., Götz, M., Hamdan, S., Komeyer, V., Kulkarni, A., Lahnakoski, J., Love, B. \& Others On Leakage in Machine Learning Pipelines. {\em ArXiv Preprint ArXiv:2311.04179}. (2023)
\bibitem{nygaard2016methods}Nygaard, V., Rødland, E. \& Hovig, E. Methods that remove batch effects while retaining group differences may lead to exaggerated confidence in downstream analyses. {\em Biostatistics}. \textbf{17}, 29-39 (2016)
\bibitem{illakiya2023automatic}Illakiya, T. \& Karthik, R. Automatic detection of Alzheimer's disease using deep learning models and neuro-imaging: current trends and future perspectives. {\em Neuroinformatics}. \textbf{21}, 339-364 (2023)
\bibitem{more2023brain}More, S., Antonopoulos, G., Hoffstaedter, F., Caspers, J., Eickhoff, S., Patil, K., Initiative, A. \& Others Brain-age prediction: A systematic comparison of machine learning workflows. {\em NeuroImage}. \textbf{270} pp. 119947 (2023)
\bibitem{wernly2021machine}Wernly, B., Mamandipoor, B., Baldia, P., Jung, C. \& Osmani, V. Machine learning predicts mortality in septic patients using only routinely available ABG variables: a multi-centre evaluation. {\em International Journal Of Medical Informatics}. \textbf{145} pp. 104312 (2021)
\bibitem{gaser2022cat}Gaser, C., Dahnke, R., Thompson, P., Kurth, F., Luders, E. \& Initiative, A. CAT–A computational anatomy toolbox for the analysis of structural MRI data. {\em Biorxiv}. pp. 2022-06 (2022)
\bibitem{snoek2021amsterdam}Snoek, L., Miesen, M., Beemsterboer, T., Van Der Leij, A., Eigenhuis, A. \& Steven Scholte, H. The Amsterdam Open MRI Collection, a set of multimodal MRI datasets for individual difference analyses. {\em Scientific Data}. \textbf{8}, 85 (2021)
\bibitem{nooner2012nki}Nooner, K., Colcombe, S., Tobe, R., Mennes, M., Benedict, M., Moreno, A., Panek, L., Brown, S., Zavitz, S., Li, Q. \& Others The NKI-Rockland sample: a model for accelerating the pace of discovery science in psychiatry. {\em Frontiers In Neuroscience}. \textbf{6} pp. 152 (2012)
\bibitem{shafto2014cambridge}Shafto, M., Tyler, L., Dixon, M., Taylor, J., Rowe, J., Cusack, R., Calder, A., Marslen-Wilson, W., Duncan, J., Dalgleish, T. \& Others The Cambridge Centre for Ageing and Neuroscience (Cam-CAN) study protocol: a cross-sectional, lifespan, multidisciplinary examination of healthy cognitive ageing. {\em BMC Neurology}. \textbf{14} pp. 1-25 (2014)
\bibitem{wei2018structural}Wei, D., Zhuang, K., Ai, L., Chen, Q., Yang, W., Liu, W., Wang, K., Sun, J. \& Qiu, J. Structural and functional brain scans from the cross-sectional Southwest University adult lifespan dataset. {\em Scientific Data}. \textbf{5}, 1-10 (2018)
\bibitem{tipping1999relevance}Tipping, M. The relevance vector machine. {\em Advances In Neural Information Processing Systems}. \textbf{12} (1999)
\bibitem{tolles2016logistic}Tolles, J. \& Meurer, W. Logistic regression: relating patient characteristics to outcomes. {\em Jama}. \textbf{316}, 533-534 (2016)
\bibitem{breiman2001random}Breiman, L. Random forests. {\em Machine Learning}. \textbf{45} pp. 5-32 (2001)
\bibitem{jack2008alzheimer}Jack Jr, C., Bernstein, M., Fox, N., Thompson, P., Alexander, G., Harvey, D., Borowski, B., Britson, P., L. Whitwell, J., Ward, C. \& Others The Alzheimer's disease neuroimaging initiative (ADNI): MRI methods. {\em Journal Of Magnetic Resonance Imaging: An Official Journal Of The International Society For Magnetic Resonance In Medicine}. \textbf{27}, 685-691 (2008)
\bibitem{yang2023predicting}Yang, Z., Cui, X. \& Song, Z. Predicting sepsis onset in ICU using machine learning models: a systematic review and meta-analysis. {\em BMC Infectious Diseases}. \textbf{23}, 635 (2023)
\bibitem{wu2021artificial}Wu, M., Du, X., Gu, R. \& Wei, J. Artificial intelligence for clinical decision support in sepsis. {\em Frontiers In Medicine}. \textbf{8} pp. 665464 (2021)
\bibitem{zhang2023machine}Zhang, Y., Xu, W., Yang, P. \& Zhang, A. Machine learning for the prediction of sepsis-related death: a systematic review and meta-analysis. {\em BMC Medical Informatics And Decision Making}. \textbf{23}, 283 (2023)
\bibitem{caspers2014studying}Caspers, S., Moebus, S., Lux, S., Pundt, N., Schütz, H., Mühleisen, T., Gras, V., Eickhoff, S., Romanzetti, S., Stöcker, T. \& Others Studying variability in human brain aging in a population-based German cohort—rationale and design of 1000BRAINS. {\em Frontiers In Aging Neuroscience}. \textbf{6} pp. 149 (2014)
\bibitem{fischl2012freesurfer}Fischl, B. FreeSurfer. {\em Neuroimage}. \textbf{62}, 774-781 (2012)
\bibitem{pollard2019eicu}Pollard, T., Johnson, A., Raffa, J., Celi, L., Badawi, O. \& Mark, R. eICU collaborative research database (version 2.0). {\em PhysioNet}. \textbf{10} pp. C2WM1R (2019)
\bibitem{pollard2018eicu}Pollard, T., Johnson, A., Raffa, J., Celi, L., Mark, R. \& Badawi, O. The eICU Collaborative Research Database, a freely available multi-center database for critical care research. {\em Scientific Data}. \textbf{5}, 1-13 (2018)
\bibitem{goldberger2000physiobank}Goldberger, A., Amaral, L., Glass, L., Hausdorff, J., Ivanov, P., Mark, R., Mietus, J., Moody, G., Peng, C. \& Stanley, H. PhysioBank, PhysioToolkit, and PhysioNet: components of a new research resource for complex physiologic signals. {\em Circulation}. \textbf{101}, e215-e220 (2000)
\bibitem{marzi2024efficacy}Marzi, C., Giannelli, M., Barucci, A., Tessa, C., Mascalchi, M. \& Diciotti, S. Efficacy of MRI data harmonization in the age of machine learning: a multicenter study across 36 datasets. {\em Scientific Data}. \textbf{11}, 115 (2024)
\bibitem{li2021impact}Li, Y., Ammari, S., Balleyguier, C., Lassau, N. \& Chouzenoux, E. Impact of preprocessing and harmonization methods on the removal of scanner effects in brain MRI radiomic features. {\em Cancers}. \textbf{13}, 3000 (2021)
\bibitem{ingalhalikar2021functional}Ingalhalikar, M., Shinde, S., Karmarkar, A., Rajan, A., Rangaprakash, D. \& Deshpande, G. Functional connectivity-based prediction of autism on site harmonized ABIDE dataset. {\em IEEE Transactions On Biomedical Engineering}. \textbf{68}, 3628-3637 (2021)
\bibitem{flint2020biological}Flint, C., Förster, K., Koser, S., Konrad, C., Zwitserlood, P., Berger, K., Hermesdorf, M., Kircher, T., Nenadic, I., Krug, A. \& Others Biological sex classification with structural MRI data shows increased misclassification in transgender women. {\em Neuropsychopharmacology}. \textbf{45}, 1758-1765 (2020)
\bibitem{lones2021avoid}Lones, M. How to avoid machine learning pitfalls: a guide for academic researchers. {\em ArXiv Preprint ArXiv:2108.02497}. (2021)
\bibitem{kapoor2023leakage}Kapoor, S. \& Narayanan, A. Leakage and the reproducibility crisis in machine-learning-based science. {\em Patterns}. \textbf{4} (2023)
\bibitem{hou2020predicting}Hou, N., Li, M., He, L., Xie, B., Wang, L., Zhang, R., Yu, Y., Sun, X., Pan, Z. \& Wang, K. Predicting 30-days mortality for MIMIC-III patients with sepsis-3: a machine learning approach using XGboost. {\em Journal Of Translational Medicine}. \textbf{18} pp. 1-14 (2020)
\bibitem{deng2022evaluating}Deng, H., Sun, M., Wang, Y., Zeng, J., Yuan, T., Li, T., Li, D., Chen, W., Zhou, P., Wang, Q. \& Others Evaluating machine learning models for sepsis prediction: A systematic review of methodologies. {\em Iscience}. \textbf{25} (2022)
\bibitem{bostami2022multi} 
Bostami, B., Espinoza, F., Horn, H., Van~Der~Naalt, J., Calhoun, V. \& Vergara, V. 
Multi-site mild traumatic brain injury classification with machine learning and harmonization. 
{\em 2022 44th Annual International Conference of the IEEE Engineering in Medicine \& Biology Society (EMBC)}. 
pp.~537--540 (2022).
\bibitem{castaldo2022framework}Castaldo, R., Brancato, V., Cavaliere, C., Trama, F., Illiano, E., Costantini, E., Ragozzino, A., Salvatore, M., Nicolai, E. \& Franzese, M. A framework of analysis to facilitate the harmonization of multicenter radiomic features in prostate cancer. {\em Journal Of Clinical Medicine}. \textbf{12}, 140 (2022)

\end{thebibliography}
\end{document}